\documentclass[11pt,a4paper]{article}

\usepackage[margin=1in]{geometry}
\usepackage{amsmath,amssymb}
\usepackage{graphicx}
\usepackage{booktabs}
\usepackage{hyperref}
\usepackage{cite}
\usepackage{microtype}
\usepackage{parskip}
\usepackage{titlesec}
\usepackage{authblk}
\usepackage{abstract}

\hypersetup{
    colorlinks=true,
    linkcolor=blue,
    citecolor=blue,
    urlcolor=blue
}

\titleformat{\section}{\large\bfseries}{\thesection}{1em}{}
\titleformat{\subsection}{\normalsize\bfseries}{\thesubsection}{1em}{}

\title{\textbf{Perforated Neural Networks for Keyword Spotting}}

\author[2]{Vishy Gopal}
\author[3]{Aris Ilias Goutis}
\author[1]{Ralph Crewe}
\author[1]{Erin Yanacek}
\author[1]{Rorry Brenner}

\affil[1]{Perforated AI}
\affil[2]{Purdue University}
\affil[3]{Renesas Electronics}

\date{}

\begin{document}

\maketitle

\begin{abstract}
Edge machine learning presents a unique set of constraints not encountered in cloud-scale
model deployment: strict memory budgets, limited compute, and non-negotiable accuracy
thresholds must all be satisfied simultaneously. Existing compression and optimization
techniques can trade one resource for another, but rarely improve both accuracy and
model size at the same time. This paper presents the application of Perforated
Backpropagation~\cite{brenner2025perforated} to keyword spotting on the Edge Impulse
platform, an experiment that won the \textit{Best Model} award at the Edge Impulse 2025
Hackathon in December 2025~\cite{edgeimpulse2025winners}. By adding artificial Dendrite Nodes to a standard convolutional neural network trained on the Edge Impulse keyword spotting tutorial pipeline, we demonstrate that dendritic models outperform traditional architectures at every level of
parameter count and at every accuracy threshold tested across 800 hyperparameter trials.
The best dendritic model achieved a test accuracy of 0.933 with only 1,500 parameters,
versus the baseline accuracy of 0.921 requiring approximately 4,000 parameters. These
results suggest that Perforated Backpropagation is a powerful addition to the edge AI
engineer's toolkit, offering simultaneous gains in both model quality and deployment
efficiency.
\end{abstract}

\section{Introduction}

Deploying machine learning models on edge devices, microcontrollers (MCUs), constrained
system-on-chips (SoCs), and other resource-limited hardware, requires satisfying two
demands that are in constant tension: high predictive accuracy and a small computational
footprint. A model that is too large will not run on the target hardware regardless of its
accuracy; a model that is too small may fit in memory but fail to meet the minimum
accuracy threshold required for a reliable product.

Keyword spotting (KWS) is one of the canonical edge ML workloads. Always-on voice
interfaces, smart home devices, and industrial monitoring systems all require a model
capable of reliably detecting a small vocabulary of trigger words under real-world noise
conditions, while running continuously on a battery-powered device~\cite{warden2018speech,
zhang2017hello}. The task is therefore a natural proving ground for any technique that
promises improvements in accuracy, model size, or both.

The artificial neuron as a computational unit has remained essentially unchanged since its
mathematical formalization in 1943~\cite{mcculloch1943logical} and its instantiation as
the perceptron in 1958~\cite{rosenblatt1958perceptron}. Backpropagation was introduced by Rumelhart, Hinton, and Williams~\cite{rumelhart1986learning}, and while activation functions, optimizers, schedulers, and large-scale architectures have evolved dramatically, the core neuron unit and the
gradient-descent learning rule have not. Modern neuroscience, however, has made
substantial progress in understanding that biological neurons are far more computationally
rich than their artificial counterparts. The most notable of which is the active, nonlinear
processing performed by \textit{dendrites}~\cite{major2013active}.

Perforated Backpropagation (PB)~\cite{brenner2025perforated} is a recently proposed
framework that updates the artificial neuron model to incorporate a form of dendritic
computation. Rather than replacing existing architectures, PB wraps them: after a network
has been trained to convergence, artificial \textit{Dendrite Nodes} are added to each
neuron through a process called \textit{perforation}. These nodes learn to correlate their output with the residual error of the associated neuron via a modified Cascade Correlation rule~\cite{fahlman1989cascade}, while gradient descent error is \textit{not} propagated through them. This ``perforation'' of the backpropagation graph keeps Dendrite Nodes architecturally outside
the network, in analogy to the role biological dendrites play relative to the neuronal
soma. The result is a plug-in that can be dropped into any PyTorch training pipeline with
minimal code changes, improving either accuracy at fixed model size or reducing model
size at fixed accuracy.

A follow-up study~\cite{brenner2025further} expanded the experimental scope of PB through
a hackathon held at the Carnegie Mellon Swartz Center in February 2025, demonstrating
gains across language modeling (BERT variants), protein sequence classification
(ProteinBERT), and image recognition (MobileNet V3), and reporting up to 90\% model
compression without accuracy loss and up to 16\% accuracy improvement.

The present paper reports the application of PB to the keyword spotting domain in the
context of the Edge Impulse 2025 Hackathon. Section~\ref{sec:background} reviews the
neuroscientific and computational foundations of Perforated Backpropagation. Section~\ref{sec:relatedwork}
situates this work in the broader context of dendritic ML and edge optimization research.
Section~\ref{sec:method} describes the experimental setup, including the Edge Impulse
platform, the keyword spotting pipeline, and the hyperparameter sweep. Section~\ref{sec:results}
presents results across 800 trials. Section~\ref{sec:conclusion} concludes.

\section{Background}
\label{sec:background}

\subsection{Active Dendrites in Biological Neurons}

The human brain remains the most capable general recognition and inference system known.
Its visual system, for instance, achieves a level of generalized object recognition that
artificial systems have only begun to approach~\cite{krizhevsky2012imagenet}. Artificial
neural networks are inspired by biological neurons, but the original artificial neuron, which remains the foundation of all modern, non-dendritic networks, is a significantly reduced model: presynaptic neurons form connections directly to the cell body, with a single connection weight as the only parameter~\cite{widrow1990thirty}. This model reflects the neuroscience of 1943 rather than modern understanding.

One key function missing from artificial neurons is the active computational role of
biological dendrites. Dendrites are not simply passive conduits for delivering presynaptic
activation to the soma; they perform complex nonlinear computations as signals travel
along them~\cite{major2013active}. In particular, dendritic branches can generate
localized N-methyl-D-aspartate (NMDA) receptor-mediated spikes. NMDA receptors are the primary excitatory neurotransmitter receptors in the brain. The NMDA receptor's current--voltage relationship has a characteristic N-shape that depends on glutamate concentration: at intermediate
concentrations, an unstable voltage threshold emerges, above which the membrane
depolarizes rapidly to a high-voltage stable state. This creates a thresholded,
all-or-none response whose threshold voltage itself varies with glutamate
concentration~\cite{major2008spatiotemporally}. The result is that a single dendritic
branch can implement nonlinear coincidence detection and feature gating, endowing a single
biological neuron with computational power that would require multiple layers of
traditional artificial neurons to replicate.

Some researchers have argued that dendrites, and not neurons, should be regarded as the
fundamental functional unit of the nervous system~\cite{branco2010single}. This
perspective motivates the addition of dendritic-like modules to artificial networks which, until now, have no analog for the functionality of dendrites.

\subsection{Cascade Correlation}

Perforated Backpropagation draws on an earlier algorithm, Cascade
Correlation~\cite{fahlman1989cascade}, which also introduces nodes that learn differently
from the primary network. In Cascade Correlation, candidate nodes are trained to maximize
their \textit{correlation} with the output error of the network while the existing weights
are frozen. The best candidate is then frozen and added to the network, after which the
output nodes resume standard learning with the new node as an additional input. Successive
cascade nodes also receive input from all previously added cascade nodes.

Cascade Correlation was originally designed for single-layer perceptron networks. PB
generalizes the cascade concept to deep networks by replacing the ``perceptron output
error'' target with the backpropagated error gradient at each individual neuron in the
deep network.

\subsection{Perforated Backpropagation}

In standard backpropagation~\cite{rumelhart1986learning} through a multilayer perceptron, the error associated with
neuron node $j$ is:
\begin{equation}
\Delta_j = g'(\mathrm{in}_j)\,\sum_i W_{i,j}\,\Delta_i
\end{equation}
where $g'(\mathrm{in}_j)$ is the derivative of the activation function applied to the
total input to neuron $j$, and $W_{i,j}$ is the weight from postsynaptic neuron $i$ to
presynaptic neuron $j$.

Perforated Backpropagation modifies this equation by splitting postsynaptic nodes into
neuron nodes (indexed $i$) and Dendrite Nodes (indexed $k$), and zeroing out the Dendrite
Node error terms during backpropagation:
\begin{equation}
\Delta_j = g'(\mathrm{in}_j)\!\left(\sum_i W_{i,j}\,\Delta_i \;+\; 0 \cdot \sum_k W_{k,j}\,\Delta_k\right)
\end{equation}

The Dendrite Node error $\Delta_k$ is computed separately using a modified Cascade
Correlation formula. During each mini-batch:
\begin{equation}
\Delta_k = \bigl(g(\mathrm{in}_k) - \bar{g}(\mathrm{in}_k)\bigr)\bigl(\Delta_i - \bar{\Delta}_i\bigr)
\end{equation}
where $g(\mathrm{in}_k)$ is the Dendrite Node activation, $\bar{g}(\mathrm{in}_k)$ is a
running average of that activation, and $\bar{\Delta}_i$ is a running average of the
error of the associated neuron node. The weight update rule for Dendrite Node input
connections is:
\begin{equation}
\frac{\delta\Delta_k}{\delta w_j} = \sigma\,(\Delta_i - \bar{\Delta}_i)\,g'(\mathrm{in}_k)
\end{equation}
where $\sigma$ is the sign of the average correlation between the Dendrite Node's output
and the associated neuron's error.

Training alternates between two phases. In the \textit{neuron phase}, the network trains
via standard gradient descent until validation performance plateaus. In the
\textit{dendrite phase}, the neuron weights are frozen, Dendrite Nodes are added to every
neuron, and they are trained using Equations~(3) and~(4) until their correlations plateau.
The best Dendrite Node per neuron is then frozen and incorporated into the forward pass,
and neuron-phase training resumes. This cycle repeats until adding further Dendrite Nodes
no longer improves validation performance. Because error is never propagated \textit{through}
Dendrite Nodes, they remain outside the gradient-descent graph. This architectural property
draws an explicit analogy to the role of biological dendrites outside the soma's
final action-potential generation.

PB is implemented as a PyTorch wrapper requiring minimal changes to existing model
code~\cite{perforatedai2024github}.

\section{Related Work}
\label{sec:relatedwork}

\paragraph{Dendritic models in machine learning.}
The idea of incorporating dendritic structures into artificial neurons dates to at least
2003~\cite{ritter2003morphological}. Early morphological perceptron work showed that
dendrite-like structures could perfectly classify any training dataset through hypercube
partitioning~\cite{sossa2014efficient}. A hardware-oriented dendritic architecture
demonstrated power-efficient inference~\cite{li2020power}. A 2021 review found that while
none of the dendritic models surveyed achieved state-of-the-art accuracy, they consistently
outperformed traditional networks at equal parameter counts~\cite{chavlis2021drawing}. PB
is distinguished from all of these by being a plug-in to existing deep architectures
rather than a replacement architecture: it requires no changes to model definitions,
training hyperparameters, or datasets. Additionally, PB is one of very few that introduces dendrite nodes to multi-layer architectures, and the only method to combine separate learning rules for neuron nodes and dendrite nodes.

\paragraph{Keyword spotting for edge deployment.}
Keyword spotting has long been a motivating application for efficient neural network
design~\cite{warden2018speech}. Early deep learning approaches demonstrated that small
convolutional and recurrent networks could achieve high accuracy on limited
vocabulary~\cite{zhang2017hello}. More recent work has explored depthwise separable
convolutions and attention mechanisms to push accuracy further while meeting MCU
constraints~\cite{banbury2021mlperftiny}. Edge Impulse provides a managed pipeline for training, evaluating, and deploying KWS models to embedded targets, making it a natural platform for comparing
optimization techniques under realistic deployment conditions~\cite{edgeimpulse2019}.

\paragraph{Model compression.}
Orthogonal compression techniques such as pruning, quantization, knowledge distillation, and
neural architecture search are widely used to reduce model size for edge
deployment~\cite{howard2019searching}. MobileNet~\cite{howard2018mobilenets} and its
successors are canonical examples of architectures designed from the ground up for
resource-constrained targets. PB differs from these in that it can both \textit{increase}
accuracy or \textit{decrease} parameter count, rather than accepting an accuracy penalty in exchange for compression.  Which it will do depends on the choices for how to integrate the method compared to the original architecture.

\paragraph{Broader PB results.}
Previous PB experiments~\cite{brenner2025perforated} demonstrated improvements on drug
toxicity prediction (TrimNet on Tox21~\cite{li2021trimnet}, 13.6\% average error
reduction), stock trend forecasting (HIST on CSI300~\cite{xu2021hist}, 8.6\% average
improvement), and ICU mortality prediction (mTAN on PhysioNet~\cite{shukla2021multitime}),
as well as model compression achieving accuracy parity with less than one-tenth the
parameters. The follow-up study~\cite{brenner2025further} extended results to BERT
variants~\cite{kenton2019bert} on NLP benchmarks, ProteinBERT on antimicrobial peptide
classification, and MobileNet V3~\cite{howard2019searching} on CIFAR-10, with deployed
inference cost reductions of up to 38$\times$ on GPU cloud hardware.

\section{Method}
\label{sec:method}

\subsection{Platform and Task}

Edge Impulse is a machine learning development platform oriented toward embedded and
edge deployment targets. Its standard \textit{Neural Network Impulse Block} provides a
configurable convolutional neural network with fully-connected network head that can be trained on audio, sensor, or image data and exported to a wide variety of MCUs and SoCs. The keyword spotting tutorial pipeline within Edge Impulse uses mel-frequency cepstral coefficient (MFCC) features extracted from short audio windows as input to this network, targeting a small vocabulary of trigger words.

\subsection{Dendritic NN Impulse Block}

To apply PB within the Edge Impulse ecosystem, we implemented a custom
\textit{Dendritic NN Impulse Block} that extends the existing NN block. The custom block
preserves full compatibility with Edge Impulse's training and deployment pipeline while
introducing the alternating neuron/dendrite training schedule described in
Section~\ref{sec:background}. This allows users access to all of the same configuration settings the original block offers, with dendritic options now included as well. All model code is open source and available at the Perforated AI GitHub repository~\cite{perforatedai2025block}.

\subsection{Hyperparameter Sweep}

We conducted a large-scale hyperparameter sweep of 800 training trials using Weights \&
Biases~\cite{wandb2020}. Both Gradient Descent (GD) dendrites, which train Dendrite Node weights with standard gradient descent, and Cascade Correlation (CC) dendrites were evaluated alongside traditional (non-dendritic) networks. The following hyperparameters were explored:

\begin{itemize}
    \item \textbf{Network architecture:} Total convolutional layers, total linear layers,
     network width, width growth mode (how channels scale across layers, 
     e.g., uniform as in $[8, 8, 8, 8]$ or doubling as in $[8, 16, 32, 64]$).
    \item \textbf{Regularization and training:} Dropout rate, Gaussian noise standard
    deviation, learning rate, early stopping patience.
    \item \textbf{Dendritic parameters:} Maximum dendrites per neuron, dendritic switch
    threshold, dendritic initialization magnitude, dendritic forward function, dendritic
    modules to perforate (linear only vs.\ linear and convolutional).
    \item \textbf{Model format:} Traditional neural network, GD dendrites, or CC dendrites.
\end{itemize}

The sweep covered a broad range of architectural sizes, from very small networks well
below the baseline parameter count to larger networks exceeding it. The original dataset is split into training, test, and validation subsets.  Test scores were recorded at the epoch of maximum validation performance (not the peak test score), consistent with standard model selection practice and with prior PB evaluations~\cite{brenner2025perforated}.

\section{Results}
\label{sec:results}

\subsection{Baseline}

The baseline Edge Impulse keyword spotting model achieves a test accuracy of 0.921 with 3,859 parameters.

\subsection{Sweep Results}

\begin{figure}[h]
    \centering
    \includegraphics[width=\columnwidth]{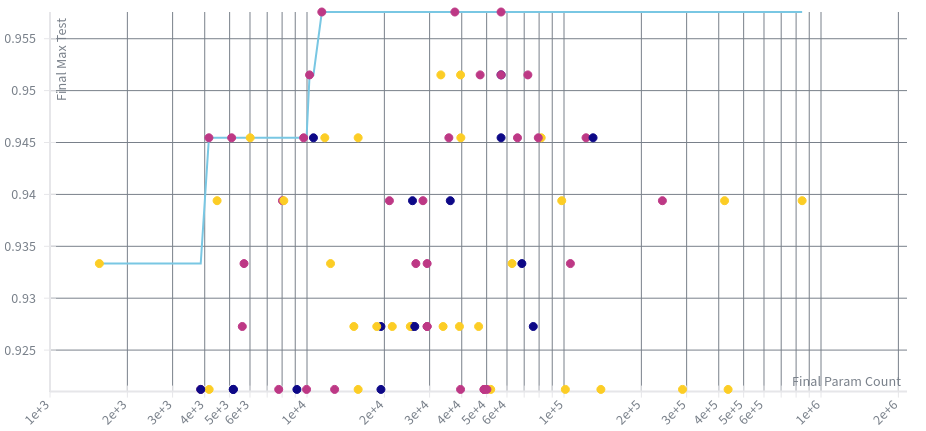}
    \caption{A focused view of the graph showing all trials that achieved above 0.920 accuracy.}
    \label{fig:sweep_zoomed}
\end{figure}

Figure~\ref{fig:sweep_zoomed} displays all 800 trial outcomes from the accompanying Weights \& Biases report~\cite{brenner2025wandbreport}. Each point represents a single trial, with
the x-axis showing total parameter count of the final deployed model and the y-axis
showing test accuracy. X-axis is logarithmic from 100-2,000,000 parameters. Blue points are traditional (non-dendritic) models, pink points are GD-dendrite models, yellow points are CC-dendrite models.

Two high-level patterns are immediately visible in the full sweep:

\begin{itemize}
    \item \textbf{At every horizontal accuracy threshold}, the leftmost point (fewest
    parameters achieving that accuracy) is a dendritic model.
    \item \textbf{At every vertical parameter budget}, the highest-accuracy point is a
    dendritic model.
\end{itemize}

\subsection{Optimal Models}

For edge deployment, the most important KPIs are model size and accuracy, and model builders must decide how to select hyperparameters to optimize for those two values.  The key findings of this experiment are:

\begin{itemize}
    \item The \textbf{smallest model} exceeding baseline accuracy is a dendritic model
    with approximately \textbf{1,556 parameters} and a test accuracy of \textbf{0.933}.
    This reduces the remaining error-rate by 16\% from the baseline while also achieving a $60\%$
    reduction in parameter count relative to the baseline.
    \item The most accurate model is a dendritic model which achieves a test accuracy of 0.958 with 11,421 parameters.  The closest traditional model has an error rate 29\% higher with a similar parameter count at 10,595.
    \item For practitioners whose primary KPI is \textbf{maximum accuracy}, dendritic
    models achieve each accuracy level with fewer parameters than any traditional model.
    \item For practitioners whose primary KPI is \textbf{smallest model size} within a
    given accuracy requirement, dendritic models always achieve higher accuracy at that
    size than traditional models.
\end{itemize}

These results demonstrate that dendritic models outperform traditional models at every point along the accuracy--efficiency tradeoff, not merely on a single experiment.

\subsection{Discussion}

The universality of the dendritic advantage across parameter counts is a stronger result
than only finding a single high-performing, potential outlier, dendritic configuration. It suggests that the mechanism underlying dendritic architectures provides a consistent signal improvement rather than one that only manifests under specific architectural conditions.

The keyword spotting domain is particularly well-suited to demonstrating this universality
because the performance requirements are binary in nature: a model either fits on the
target hardware or it does not, and it either meets the accuracy threshold or it does not.
The fact that dendritic models dominate at every intersection of these two axes means that
any edge ML engineer constrained along either dimension would benefit from building with dendrites.

This result also extends the PB track record into the audio and edge inference domain from the domains previously reported~\cite{brenner2025perforated, brenner2025further}, graph neural networks for
chemistry, stock forecasting, time-series medical data, NLP, protein biology, and image
classification. The consistency of results across these diverse settings strengthens the case that dendritic architectures and Perforated Backpropagation captures a general property of neural network learning rather than one specific to any particular domain or architecture family.

\begin{table}[h]
\centering
\caption{Comparison of key models from the 800-trial hyperparameter sweep.}
\label{tab:results}
\begin{tabular}{lccc}
\toprule
\textbf{Model} & \textbf{Type} & \textbf{Parameters} & \textbf{Error Rate} \\
\midrule
Baseline (traditional NN) & Traditional & 3,859 & 0.079 \\
Smallest model exceeding baseline & Dendritic & \textbf{1,556} & 0.067 \\
Most accurate model & Dendritic & 11,421 & \textbf{0.042} \\
\midrule
\textbf{Optimal change} & & \textbf{$-$60\%} & \textbf{$-$46\%} \\
\bottomrule
\end{tabular}
\end{table}

\section{Conclusion}
\label{sec:conclusion}

We have presented the application of dendritic architectures and Perforated Backpropagation to keyword spotting on the Edge Impulse platform, demonstrating that artificial dendrite nodes consistently outperform traditional architectures across all operating points of the
accuracy--efficiency trade-off. Across 800 hyperparameter trials, dendritic models
achieved higher accuracy than traditional models at any given parameter count, and
achieved any given accuracy level with fewer parameters. The best single dendritic model
simultaneously reduced error rate by 16\% while also reducing parameter count by
60\% relative to the baseline, earning the Best Model award at the Edge Impulse 2025
Hackathon~\cite{edgeimpulse2025winners}.

These results add to a growing body of evidence~\cite{brenner2025perforated,
brenner2025further} that dendritic architectures and Perforated Backpropagation are a general-purpose improvement to deep neural network training. The technique is motivated by modern neuroscience
understanding of active dendritic computation~\cite{major2013active, branco2010single},
implemented in a manner compatible with any PyTorch model, and now validated across
chemistry, finance, medicine, natural language processing, protein biology, image
recognition, and audio keyword spotting.

Edge AI represents one of the most demanding deployment environments in machine
learning. The simultaneous gains in accuracy and efficiency demonstrated here suggest that artificial dendrites may be especially valuable for this setting, where every kilobyte and every percentage point of accuracy matter. Future work will explore application to additional audio architectures and
embedded deployment benchmarks, as well as integration with hardware-aware neural
architecture search.

The Dendritic NN Impulse Block and all associated code are publicly available at~\cite{perforatedai2025block}.

\bibliographystyle{plain}
\bibliography{pb_edge_impulse}

\end{document}